\documentclass{bmvc2k}
\usepackage{amsmath}
\usepackage{amssymb}
\usepackage{graphicx}
\usepackage{multirow}
\usepackage{booktabs}
\usepackage{algorithm}
\usepackage{algpseudocode}
\usepackage{caption}
\usepackage{subcaption}

\title{Video Dataset Condensation with Diffusion Models}

\addauthor{Zhe Li}{zhe.li@fau.de}{1}
\addauthor{Hadrien Reynaud}{hadrien.reynaud19@imperial.ac.uk}{2}
\addauthor{Mischa Dombrowski}{mischa.dombrowski@fau.de}{1}
\addauthor{Sarah Cechnicka}{sarah.cechnicka18@imperial.ac.uk}{2}
\addauthor{Franciskus Xaverius Erick}{franciskus.erick@fau.de}{1}
\addauthor{Bernhard Kainz}{b.kainz@imperial.ac.uk}{1,2}

\addinstitution{
 Department AIBE\\
 FAU Erlangen-Nürnberg\\
 Erlangen, Germany
}
\addinstitution{
 Department of Computing\\
 Imperial College London\\
 London, UK
}

\runninghead{Z Li, Reynaud, Dombrowski, Cechnicka, Erick, Kainz}{Video DD}

\def\etal{\emph{et al}\bmvaOneDot}

\begin{document}

\maketitle

\begin{abstract}
In recent years, the rapid expansion of dataset sizes and the increasing complexity of deep learning models have significantly escalated the demand for computational resources, both for data storage and model training. Dataset distillation has emerged as a promising solution to address this challenge by generating a compact synthetic dataset that retains the essential information from a large real dataset. However, existing methods often suffer from limited performance, particularly in the video domain.  
In this paper, we focus on video dataset distillation. We begin by employing a video diffusion model to generate synthetic videos. Since the videos are generated only once, this significantly reduces computational costs. Next, we introduce the Video Spatio-Temporal U-Net (VST-UNet), a model designed to select a diverse and informative subset of videos that effectively captures the characteristics of the original dataset. 
To further optimize computational efficiency, we explore a training-free clustering algorithm, Temporal-Aware Cluster-based Distillation (TAC-DT), to select representative videos without requiring additional training overhead. We validate the effectiveness of our approach through extensive experiments on four benchmark datasets, demonstrating performance improvements of up to \(10.61\%\) over the state-of-the-art. Our method consistently outperforms existing approaches across all datasets, establishing a new benchmark for video dataset distillation.
\end{abstract}

\section{Introduction}
\label{sec:intro}

The rapid increase in images and videos uploaded to social media, coupled with the growing complexity of deep learning models, has led to significant challenges in data storage and computational costs for training large-scale models on expanding datasets. Dataset distillation has emerged as a promising solution to address these challenges by generating a compact synthetic dataset that allows models to achieve comparable performance to those trained on the full original dataset.
While existing research~\cite{zhao2020dataset,zhao2023DM,cazenavette2022dataset,wang2022cafe,liu2022dataset,sajedi2023datadam,zhu2023rethinking,du2023minimizing} has achieved significant progress in image dataset distillation, relatively few studies have focused on video datasets. The additional temporal dimension in videos introduces greater data volume, redundancy, and computational complexity, making efficient distillation more challenging.  
Matching-based approaches~\cite{wang2024dancing}, when extended to videos, first distill representative frames and then incorporate dynamic motion information. These methods face a fundamental trade-off: retaining more frames enhances the quality of representation but also significantly increases computational costs, and the overall performance on distilled video datasets remains suboptimal.
Additionally, video dataset distillation suffers from a scalability issue. The computational time grows exponentially as the number of synthetic videos increases, thereby limiting practical applications for large-scale datasets. These limitations emphasize the need for more efficient, scalable, and structured distillation methods tailored to video-based tasks.

To overcome these challenges, we exploit a class-conditional video diffusion model Latte~\cite{ma2024latte} to effectively capture the data distribution of the training video dataset. 
A major advantage of this method is the generation time has linear scalability with respect to the number of synthetic videos, making it more practical for large-scale datasets. 
To ensure that the synthetic dataset effectively represents the data distribution, we first generate a large pool of synthetic videos and then apply various clustering techniques to select a subset of representative videos. 
Since the large number of synthetic videos is generated only once and stored, subsequent processes can focus solely on selection. As a result, the computational cost remains stable even as the videos-per-class (VPC) value increases.
The selection process is designed to maximize diversity in the final distilled dataset, thereby preserving essential motion characteristics while minimizing redundancy.
Drawing inspiration from video summarization techniques~\cite{zhou2018deep,liu2022video,son2024csta,huynh2023cluster}, which aim to select keyframes that best represent an entire video, we introduce Video Spatio-Temporal U-Net (VST-UNet) as a video selection model for synthetic datasets. By leveraging both spatial and temporal information, VST-UNet effectively identifies representative videos that best characterize the dataset. To ensure a well-balanced selection, we incorporate diversity and representativeness losses, encouraging a more comprehensive and informative subset of videos.
To further reduce computational costs, we propose a training-free clustering algorithm, Temporal-Aware Cluster-based Distillation (TAC-DT), which hierarchically constructs clusters using the BIRCH algorithm~\cite{zhang1996birch,huynh2023cluster}. This approach captures both global and local relationships between videos, allowing for efficient selection without requiring additional model training.

After selecting representative videos, we train action recognition classifiers on the distilled subset and evaluate their performance on the full real test set. Our method consistently outperforms state-of-the-art approaches by up to $10.61\%$ across various numbers of videos per class (VPC). Notably, on the miniUCF dataset, setting the VPC to 
$10$ achieves performance comparable to training on the full real dataset. This result demonstrates the effectiveness of our approach in selecting representative samples, enabling efficient model training with significantly fewer data.
Our contributions are as follows:
\begin{enumerate}
  \item We propose a novel video dataset distillation framework that leverages a diffusion model to generate a compact synthetic dataset while preserving essential information from large real datasets. This approach maintains strong representational fidelity while minimizing computational cost.

  \item We introduce two video selection strategies: (i) VST-UNet (Sec 3.2), a 4D U-Net architecture that integrates both spatial and temporal features, allowing for video selection based on learned probability distributions. (ii) TAC-DT (Sec 3.3), a training-free clustering algorithm that operates on video embeddings to efficiently identify representative videos, ensuring a well-structured and informative distilled dataset.  

  \item We demonstrate the effectiveness of our approach through extensive experiments on four datasets, achieving up to a \(10.61\%\) improvement over the current state-of-the-art approaches.
\end{enumerate}

\section{Related Work}

Dataset distillation was initially proposed alongside model selection techniques~\cite{wang2018dataset, such2020generative} for image datasets. Subsequently, researchers introduced matching-based approaches aiming to minimize the distance between real and synthetic images during training. Notable examples include Dataset Condensation with Gradient Matching (DC)~\cite{zhao2020dataset,zhao2021dataset,zhang2023accelerating}, Distribution Matching (DM)~\cite{zhao2023DM,zhao2023improved}, Matching Training Trajectories (MTT)~\cite{cazenavette2022dataset}, Sequential Subset Matching~\cite{du2024sequential}, and feature alignment in convolutional networks~\cite{wang2022cafe,sajedi2023datadam}.
Due to the presence of high-frequency noise in pixel space, some methods~\cite{zhao2022synthesizing, cazenavette2023generalizing} synthesize images directly in the latent space using pretrained generative adversarial networks (GANs). Other studies have tackled alternative phases of dataset distillation, including defining optimal distillation spaces~\cite{liu2023fewshot}, clustering methods for real-image selection~\cite{liu2023dream}, and novel matching metrics based on mutual information~\cite{shang2024mim4dd}.
Recently, diffusion models have emerged as a promising approach within dataset distillation frameworks. Su~\etal~\cite{su2024d} proposed a latent diffusion model that trains directly on prototypes rather than employing a matching process, significantly reducing computational complexity while maintaining high-quality image outputs.

While various methods for image distillation have been well-established, video distillation remains relatively underexplored. Wang~\etal~\cite{wang2024dancing} introduced a two-stage framework that initially applies existing image-based matching methods, subsequently integrating temporal dynamics to encapsulate motion information. In contrast, our proposed approach directly operates on videos using a diffusion model without an intermediate image-based stage. This enables our method to overcome inherent limitations of matching-based methods, such as the necessity for strict temporal consistency and uniform video lengths, thus providing greater adaptability to real-world datasets.

\section{Method}
\label{sec:method}

\begin{figure*}
\begin{center}
\includegraphics[width=1\linewidth]{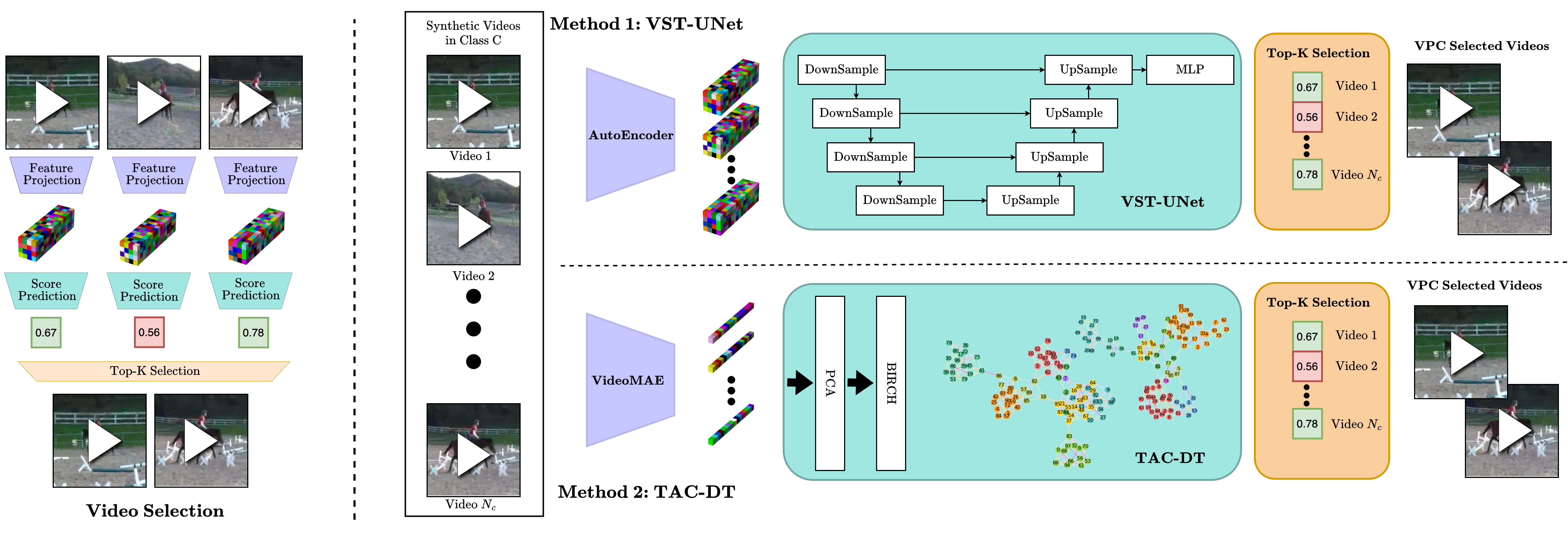}
\end{center}
   \caption{Overview of Our Framework. The left illustrates the video selection process. On the right, VST-UNet (top) extracts video latent features via an Autoencoder for selection, while TAC-DT (bottom) selects representative videos without training using VideoMAE embeddings~\cite{wang2023videomae}.}
\label{fig:overviewframework}
\end{figure*}

Dataset distillation aims to condense essential information from a large-scale real dataset $\mathcal{T}=\{(x_i, y_i)\}_{i=1}^{N}$ into a significantly smaller synthetic dataset $\mathcal{S}=\{(s_i, y_i)\}_{i=1}^{N_s}$. Each video $x_i \in \mathbb{R}^{T \times 3 \times H \times W}$ has an associated class label $y_i \in \{0, 1, 2, \dots, C\}$. The total number of synthetic videos is $N_s = \text{VPC (Videos Per Class)} \times C$ and $N_s \ll N$. The goal is to enable models trained on this condensed dataset to achieve performance comparable to models trained on the original, larger dataset, while significantly reducing the data volume required.
To effectively capture the distribution of the real dataset $\mathcal{T}$, we fine-tune a video diffusion model, Latte~\cite{ma2024latte}, generating a comprehensive set of synthetic videos $\mathcal{S}_g = \mathcal{F}_{\theta}(\mathcal{T})$. Here, $\mathcal{S}_g \in \mathbb{R}^{N_g \times T \times 3 \times H \times W}$ represents the synthetic video set, with $N_g = N_c \times C$, where $N_c$ denotes the number of synthetic videos generated per class. Once generated, these synthetic videos are stored and serve as input for a crucial selection phase.
Specifically, we introduce two novel video selection methods: Video Spatio-Temporal U-Net (VST-UNet, Section~\ref{sec:vstunet}) and Temporal-Aware Cluster-based Distillation (TAC-DT, Section~\ref{sec:tacdt}).
These approaches identify highly representative subsets from the initially synthesized videos, resulting in the final distilled dataset. Figure~\ref{fig:overviewframework} provides an overview of our proposed framework.

\subsection{VST-UNet}
\label{sec:vstunet}

The objective of VST-UNet is to identify the most representative videos within each class in the synthetic dataset \(S_g\). To achieve this, we consider both the spatiotemporal information within each video and the relationships among videos within the same class, as they often share common motion patterns and structural similarities. 
Drawing inspiration from video summarization techniques~\cite{zhou2018deep,liu2022video}, which focus on selecting key frames from individual videos, we propose a 4D Spatio-Temporal U-Net to enable selection at the video level from an entire batch as in Figure~\ref{fig:overviewframework} \textit{Video Selection}. The U-Net architecture has been widely recognized for its effectiveness in capturing semantic spatial features. By treating all videos in a class as a single structured input, our approach ensures that temporal information is not only extracted from continuous frames within individual videos but also from the relative relationships among all videos within the class.

Since the representativeness of individual videos is unknown, we adopt an unsupervised learning strategy to train VST-UNet, enabling class-wise video selection without relying on ground-truth annotations. During training, each input batch $V_c \in \mathbb{R}^{N_c \times T \times 3 \times H \times W}$ is composed solely of videos from the same class, rather than mixing videos from different classes. Here, $N_c$ denotes the number of synthetic videos for class $c$ in $S_g$. This design allows VST-UNet to effectively learn both intra-video and inter-video dependencies, enabling more informed and discriminative video selection.
To enhance computational efficiency, we project all videos into a latent space using a pretrained Autoencoder. This transformation reduces the video dimensions from \(V \in \mathbb{R}^{T\times 3 \times H \times W} \) to \(\hat{V} \in \mathbb{R}^{T\times 4 \times h \times w} \), where \(h = \frac{H}{s}, \, w = \frac{W}{s}, \, s \in \mathbb{Z}^+ \). This step significantly reduces memory requirements and computational costs.  
We input the latent video features $\hat{V}_c \in \mathbb{R}^{N_c \times T \times 4 \times h \times w}$ into the VST-UNet selection model, which outputs both the probabilities $p_c$ and the feature embeddings $f_c$ from its penultimate layer. A small subset of videos with the highest probabilities is chosen in each epoch as the representative subset $\mathcal{S}_c$ for each class. These selected videos are then used in the loss computation during training.
We introduce two reward-based losses, inspired by reinforcement learning, to enhance selection effectiveness: diversity loss and representativeness loss. The diversity loss \( L_{div} \) measures the dissimilarity among the selected videos within each class, as in Eq.~\ref{lossdiv}.

\begin{equation}\label{lossdiv}
\begin{split}
    L_{div} = \frac{1}{|\mathcal{S}_c|(|\mathcal{S}_c|-1)} \sum_{t\in \mathcal{S}_c} \sum_{i\in \mathcal{S}_c \atop{i\neq t}} d(x_t, x_i)
\end{split}
\end{equation}

where \( d(\cdot, \cdot) \) denotes the cosine similarity between two video embeddings. By minimizing the overall cosine similarity, we encourage the selected videos to be as dissimilar as possible, promoting a wide distribution across the feature space and ensuring that the selected samples are well-spread within each class.
The representativeness loss \(L_{rep}\) encourages the selected videos to effectively represent all synthetic videos by minimizing the distance between the selected videos and all other videos in each class, while maintaining a maximum distance between them as in Eq.~\ref{lossrep}. Informally, this can be thought of as extracting the support vectors of the high-dimensional video distribution. We use the mean square error to quantify the distance between two videos.

\begin{algorithm}[tb]
\caption{Our proposed algorithm for video distillation
}\label{algo:unet3d}

\begin{algorithmic}[1]
\State \textbf{Input}: synthetic videos $\mathcal{S}_g$, VPC;
\State Initialize the selection VST-UNet model \(M_{VST}\) 
\For{ \(E_s\) epochs}
    \For{class \(c\)}
        \State \(V_c \leftarrow \mathcal{S}_g\)
        \State \(\hat{V}_c = \) AutoEncoder(\(V_c\))
        \State \(f_c,\ p_c = M_{VST}(\hat{V}_c)\)
        \State \(L = L_{ce} + \lambda_dL_{div} + \lambda_rL_{rep}\)
    \EndFor
\EndFor
\For{class \(c\)}
\State \(f_c,\ p_c = M_{VST}(\hat{V}_c)\)
\State \(S_c = V_c\left[argsort(p_c)[:\text{VPC}]\right]\)
\EndFor
\State Train classifiers on the selected synthetic videos \(S\).
\end{algorithmic}
\end{algorithm}

\begin{equation}\label{lossrep}
\begin{split}
    L_{rep} = \text{exp}\left(-\frac{1}{N_c} \sum_{t=1}^{N_c }\underset{i\in \mathcal{S}_c}{\text{min}} \parallel x_t - x_i \parallel_2 \right)
\end{split}
\end{equation}

\begin{equation}\label{lossall}
\begin{split}
    L = L_{ce} + \lambda_dL_{div} + \lambda_rL_{rep}
\end{split}
\end{equation}

The cross-entropy loss \( L_{ce} \) is also used as the class-dependent objective. The final loss is the sum of all three losses as in Eq~\ref{lossall}.
Once trained, the final selected videos are determined in inference for the downstream task. 
Algorithm~\ref{algo:unet3d} outlines the process of training a VST-UNet model to select synthetic videos.


\subsection{TAC-DT}
\label{sec:tacdt}

To further reduce training costs beyond model training, we investigate the use of training-free clustering algorithms for video selection. Inspired by the training-free video summarization method TAC-SUM~\cite{huynh2023cluster}, we introduce Temporal-Aware Cluster-based Distillation (TAC-DT), an approach that leverages temporal relationships among videos within a class to organize video clusters into temporally coherent segments.

In this approach, each video is treated as a single entity and projected into a 1D embedding \( f_i \in \mathbb{R}^{D} \) using a pretrained VideoMAE~\cite{wang2023videomae} action recognition classifier. We then aggregate the embeddings of all videos belonging to the same class into a collective representation \( f_c \in \mathbb{R}^{N_c \times D} \), analogous to how frame features are embedded to represent a video, where \( N_c \) denotes the number of videos in class \( c \). To enhance computational efficiency, these video embeddings are further dimensionally reduced using Principal Component Analysis (PCA).
To capture both global and local relationships between videos, we construct clusters using the BIRCH (Balanced Iterative Reducing and Clustering using Hierarchies) algorithm~\cite{zhang1996birch}, which is a hierarchical clustering technique designed to efficiently handle large datasets by incrementally constructing a tree-based representation of data. 
Representative videos are then selected from each class, and this process is iteratively applied across all classes.

Our TAC-DT applies clustering at the inter-video level, grouping videos with similar motion characteristics within a class. This enables TAC-DT to preserve essential motion dynamics while ensuring diverse selection across the dataset. The hierarchical clustering structure of BIRCH further enhances scalability and adaptability, allowing TAC-DT to efficiently process large synthetic video datasets without requiring extensive computational resources. By integrating BIRCH algorithm, our approach achieves a robust balance between representativeness, diversity, and computational efficiency, making it an effective solution for training-free video dataset distillation.

\section{Experiments}

\begin{table}[tb]\setlength{\tabcolsep}{8pt}
  \begin{center}
  \begin{tabular}{lrrrrr}
    \toprule
    Dataset  & C & Total & Training & Val & Generated \\
    \midrule
     UCF101 & 101 & 13,320 & 9,537 & 3,783 & 10,100  \\
     miniUCF & 50 & 6,523 & 4,662 & 1,861 & 5,000\\
     HMDB51 & 51 & 6,766 & 5,236 & 1,530 & 5,100 \\
     Kinetics 400 & 400 & 254,379 & 234,619 & 19,760 & 40,000 \\
     SSv2 & 174 & 220,847 & 168,913 & 24,777 & 17,400 \\
    \bottomrule
  \end{tabular}
  \end{center}
  \caption{Datasets introduction. 'SSv2' is Something Something v2 dataset.}
  \label{datasetstat}
\end{table}

We evaluate our approach on \(4\) datasets: 
the UCF101~\cite{soomro2012ucf101} dataset and its subset, MiniUCF, which consists of 50 classes of videos with higher class-wise test accuracy, the HMDB51 \cite{kuehne2011hmdb} dataset, the Kinetics 400~\cite{kay2017kinetics} dataset, and the Something Something v2~\cite{goyal2017something} dataset. The dataset statistics are provided in Table~\ref{datasetstat}. 
Each dataset consists of short video clips, with each clip containing a single action, making them well-suited for action recognition tasks.
We evaluate our approach using three metrics: (i) Top-1 test accuracy on the MiniUCF, UCF101, and HMDB51 datasets using the real test sets; (ii) Top-1 and Top-5 test accuracy on the Kinetics400 and Something-Something V2 datasets; and (iii) AUC score and F1 score across all datasets.

We fine-tune the diffusion model weights using the pretrained Latte~\cite{ma2024latte} model. After training, we generate \(N_c=100\) synthetic videos per class, with the total number of synthetic videos detailed in Table~\ref{datasetstat}.
During the video selection phase, an AutoEncoder extracts latent features of size \(16 \times 4 \times 32 \times 32\) from input videos with original dimensions of \(16 \times 3 \times 256 \times 256\). 
The feature dimension \(D\), obtained from the pretrained VideoMAE, is \(1408\).
The hyperparameters in the loss function (Eq.~\ref{lossall}) are configured as follows:  
When VPC \(=1\): \(\lambda_r=0.1\) and \(\lambda_d=0\), as diversity loss is not applicable when selecting only one video per class; when VPC \(=5\) or \(10\): \(\lambda_r=1\) and \(\lambda_d=0.1\), ensuring a balance between representativeness and diversity in selection.  
The classifier is C3D~\cite{tran2015learning} in all experiments. To ensure the robustness of our approach, we repeat both the selection process and classification task \(5\) times and report the average and standard deviation across all runs. The video selection process is updated in each run, mitigating potential biases and ensuring fair evaluation. 

\begin{table}[tb]
\begin{center}
\resizebox{0.99\textwidth}{!}{%
  \begin{tabular}{lcccccccc}
    \toprule
    Datasets & \multicolumn{2}{c}{MiniUCF} & \multicolumn{2}{c}{HMDB51} & \multicolumn{2}{c}{Kinetics400(top5)} & \multicolumn{2}{c}{SSv2(top5)}\\
    \midrule
    VPC & 1 & 5 & 1 & 5 & 1 & 5 & 1 & 5 \\
    \midrule
    Full real & \multicolumn{2}{c}{60.14$\pm{0.95}$} & \multicolumn{2}{c}{36.24$\pm{0.83}$} & \multicolumn{2}{c}{34.6$\pm{0.5}$} & \multicolumn{2}{c}{29.0$\pm{0.6}$} \\
    Full syn (100) & \multicolumn{2}{c}{56.22$\pm{0.44}$} & \multicolumn{2}{c}{31.05$\pm{0.30}$} & \multicolumn{2}{c}{-} & \multicolumn{2}{c}{14.56$\pm{0.15}$}\\
    \midrule
    DM~\cite{zhao2023DM,wang2024dancing} & 15.3$\pm{1.1}$ &  25.7$\pm{0.2}$ & 6.1$\pm{0.2}$ & 8.0$\pm{0.2}$ &  6.3$\pm{0.0}$ & 9.1$\pm{0.9}$ & 3.6$\pm{0.0}$ & 4.1$\pm{0.0}$ \\
    MTT~\cite{cazenavette2022dataset,wang2024dancing} & 19.0$\pm{0.1}$ &  28.4$\pm{0.7}$ & 6.6$\pm{0.5}$ & 8.4$\pm{0.6}$ & 3.8$\pm{0.2}$ & 9.1$\pm{0.3}$ & 3.9$\pm{0.1}$ & 6.3$\pm{0.3}$ \\
    FRePo~\cite{zhou2022dataset,wang2024dancing} & 20.3$\pm{0.5}$ & 30.2$\pm{1.7}$ & 7.2$\pm{0.8}$ & 9.6$\pm{0.7}$ & -& - &- &- \\
    DM+VD~\cite{wang2024dancing} & 17.5$\pm{0.1}$ & 27.2$\pm{0.4}$ & 6.0$\pm{0.4}$ & 8.2$\pm{0.1}$ & 6.3$\pm{0.2}$ & 7.0$\pm{0.1}$ & 4.0$\pm{0.1}$ & 3.8$\pm{0.1}$\\
    MTT+VD~\cite{wang2024dancing} & 23.3$\pm{0.6}$ &  28.3$\pm{0.0}$ &  6.5$\pm{0.1}$ & 8.9$\pm{0.6}$ & 6.3$\pm{0.1}$ & 11.5$\pm{0.5}$ & 5.5$\pm{0.1}$ & \textbf{8.3\(\pm{0.2}\)} \\
    FRePo+VD~\cite{wang2024dancing} & 22.0$\pm{1.0}$ &  31.2$\pm{0.7}$  & 8.6$\pm{0.5}$ & 10.3$\pm{0.6}$ & -& - &- &- \\
    \midrule
    \midrule
    VST-UNet (ours) & \textbf{26.28\(\pm{1.07}\)} & \textbf{41.81$\pm{0.99}$} & 13.59$\pm{0.33}$ & 18.05$\pm{0.75}$ &\textbf{6.67$\pm{0.11}$} & \textbf{11.82$\pm{0.10}$} & 5.25$\pm{0.16}$ & 8.22$\pm{0.14}$ \\
    TAC-DT (ours) & 25.20$\pm{0.41}$ & 41.47$\pm{0.32}$ & \textbf{13.82$\pm{0.33}$} & \textbf{19.62$\pm{0.28}$} & 6.50\(\pm{0.19}\) & 11.72\(\pm{0.35}\) & \textbf{5.54\(\pm{0.19}\)} & 8.06\(\pm{0.17}\)  \\
    \bottomrule
  \end{tabular}
  }
\end{center}
  \caption{Comparison of our approaches with the state-of-the-art on all 4 datasets. We report top-1 test accuracy for the MiniUCF and HMDB51 datasets and top-5 accuracy for the Kinetics400 and Something Something v2 (SSv2) datasets.}
  \label{comparisonsotaall}
\end{table}

\subsection{Comparison with the state-of-the-art}

\begin{figure}
\begin{center}
\includegraphics[width=1\linewidth]{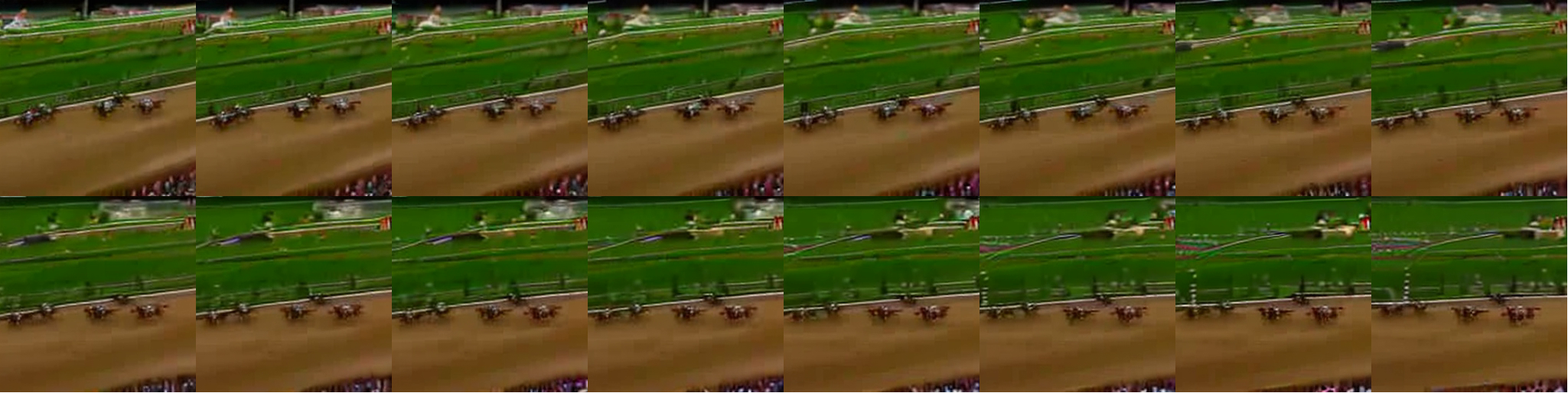}
\end{center}
   \caption{The \(16\) consecutive frames from a synthetic video of the class \textit{horserace} in the UCF101 dataset.}
\label{fig:syntheticsample}
\end{figure}

Table~\ref{comparisonsotaall} compares our approaches with state-of-the-art methods across \(4\) distilled synthetic datasets. 
We include results using the full dataset as two upper bounds: \textbf{Full real}, training on the original real dataset; \textbf{Full syn}, training on the entire synthetic dataset (\(S_g\), VPC \(= 100\)) without selection.  
To construct a small distilled dataset, we select videos from the generated synthetic dataset using our proposed selection methods, VST-UNet and TAC-DT, at varying levels of videos per class (VPC), specifically $\text{VPC} = 1$, $5$, and $10$. This results in final distilled datasets containing $\text{VPC} \times C$ videos, where $C$ is the number of classes.
Our approach consistently outperforms existing methods across different settings. On the MiniUCF dataset, we observe performance gains of $2.98\%$ ($26.28$ vs. $23.3$) when $\text{VPC} = 1$, and $10.61\%$ ($41.81$ vs. $31.2$) when $\text{VPC} = 5$. Similarly, on HMDB51, our method achieves improvements of $5.22\%$ ($13.82$ vs. $8.6$) at $\text{VPC} = 1$, and $9.32\%$ ($19.62$ vs. $10.3$) at $\text{VPC} = 5$.
These results demonstrate that our selection strategies effectively retain critical semantic information while significantly reducing dataset size, enabling efficient and high-quality video dataset distillation.
Figure~\ref{fig:syntheticsample} shows \(16\) consecutive frames from a synthetic video in the UCF101 dataset, illustrating the high visual quality and realism of generated samples.

\subsection{Ablation study}

To evaluate the generalizability of our framework, we run experiments on the full UCF101 dataset, rather than on the MiniUCF dataset, which only includes high-performing classes. Table~\ref{comparisonsotaucf101} presents the results, highlighting the robustness of our approach across a more diverse set of action categories.
Despite the inclusion of \(51\) additional classes with lower class-wise test accuracy, our framework maintains performance comparable to the state-of-the-art on MiniUCF. For instance, when VPC \(=5\), we achieve \(30.43\%\) test accuracy, which is on par with the \(31.2\%\) observed on MiniUCF. 
These results validate the effectiveness and scalability of our method, demonstrating that it can generalize beyond a curated subset of well-performing classes to a full-scale dataset with diverse motion complexities.

\begin{table}[tb]\setlength{\tabcolsep}{8pt}
\begin{center}
  \resizebox{0.6\textwidth}{!}{%
  \begin{tabular}{lccc}
    \toprule
    VPC & 1 & 5 & 10 \\
    \midrule
    Full real & \multicolumn{3}{c}{50.66\(\pm{0.81}\)} \\
    Full syn & \multicolumn{3}{c}{51.84$\pm{0.55}$} \\
    \midrule
    Random & 14.21$\pm{0.82}$ & 30.12$\pm{0.75}$ & \textbf{38.10$\pm{1.04}$} \\
    CSTA & 14.89\(\pm{0.88}\) & 29.47\(\pm{0.83}\) & 37.16\(\pm{0.81}\) \\
    VST-UNet & 16.29\(\pm{0.76}\) & 29.07\(\pm{0.39}\) & 36.61\(\pm{0.99}\) \\
    TAC-DT & \textbf{16.40\(\pm{0.55}\)} & \textbf{30.43\(\pm{0.23}\)} & 37.67\(\pm{0.61}\) \\
    \bottomrule
  \end{tabular}
  }
\end{center}
  \caption{Results from different clustering methods on the UCF101 dataset.}
  \label{comparisonsotaucf101}
\end{table}

\begin{table}[tb]
\begin{center}
  \resizebox{0.99\textwidth}{!}{%
  \begin{tabular}{llcccccccccccc}
    \toprule
     && \multicolumn{3}{c}{MiniUCF} & \multicolumn{3}{c}{HMDB51} \\
    \midrule
    & Methods & ACC & AUC & F1 Score & ACC & AUC & F1 Score \\
    \midrule
    \multirow{6}{*}{VPC=1} & Random & 23.42$\pm{1.23}$ & 77.73$\pm{0.91}$ & 21.29$\pm{1.14}$ & 11.22$\pm{0.81}$ & 64.78$\pm{1.37}$ & 9.69$\pm{0.84}$ \\
   
    & MAE-Score & 23.21$\pm{0.29}$ & 78.80$\pm{0.38}$ & 22.01$\pm{0.22}$ & 10.72$\pm{0.89}$ & 64.93$\pm{0.58}$ & 10.05$\pm{0.95}$ \\
    & CSTA & 23.80$\pm{0.79}$ & 78.35$\pm{0.90}$ & 21.15$\pm{0.91}$ & 11.92$\pm{2.05}$ & 64.37$\pm{1.93}$ & 10.58$\pm{2.15}$ \\
    & DM-MMD & 23.67$\pm{0.45}$ & 78.83$\pm{0.51}$ &  21.85$\pm{0.40}$ & 11.97$\pm{0.93}$ & 64.97$\pm{1.02}$ & 10.86$\pm{0.98}$ \\
    & VST-UNet (ours) & \textbf{26.28$\pm{1.07}$} & \textbf{80.30$\pm{0.45}$} & \textbf{24.68$\pm{0.36}$} & 13.59$\pm{0.26}$ & 65.53$\pm{0.45}$ & 11.80$\pm{0.37}$ \\
  
    & TAC-DT (ours) & 25.20\(\pm{0.41}\) & 79.94\(\pm{0.43}\) & 23.59\(\pm{0.49}\) & \textbf{13.82\(\pm{0.33}\)} & \textbf{66.83\(\pm{0.19}\)} & \textbf{12.23\(\pm{0.41}\)} \\
    \midrule
    \multirow{6}{*}{VPC=5} & Random & 38.00$\pm{1.07}$ & 87.31$\pm{0.47}$ & 35.52$\pm{0.96}$ & 18.81$\pm{0.79}$ & 71.69$\pm{0.42}$ & 17.23$\pm{0.92}$  \\
   
    & MAE-Score &  40.80$\pm{0.59}$ & 87.59$\pm{0.24}$ & 38.52$\pm{0.55}$ & 18.86$\pm{0.68}$ & 70.67$\pm{0.47}$ & 17.35$\pm{0.91}$ \\
    & CSTA & 41.77$\pm{1.15}$ & 88.51$\pm{0.34}$ & 39.93$\pm{0.85}$ & 19.57$\pm{0.89}$ & \textbf{72.85$\pm{0.76}$} & 18.42$\pm{0.80}$ \\
    & DM-MMD & 40.25$\pm{0.77}$ & 87.48$\pm{0.48}$ &  38.31$\pm{0.61}$ & 19.07$\pm{0.81}$ & 71.82$\pm{0.59}$ & 17.49$\pm{0.95}$ \\
    & VST-UNet (ours) & \textbf{41.81\(\pm{0.99}\)} & 88.45$\pm{0.54}$ & \textbf{40.07$\pm{0.95}$} & 18.05$\pm{0.75}$ & 71.87$\pm{0.78}$ & 17.15$\pm{0.80}$ \\
   
    & TAC-DT (ours) & 41.47\(\pm{0.32}\) & \textbf{88.73\(\pm{0.12}\)} & 39.81\(\pm{0.47}\) & \textbf{19.62\(\pm{0.28}\)} & 72.37\(\pm{0.11}\) & \textbf{18.61\(\pm{0.24}\)} \\
    \midrule
    \multirow{6}{*}{VPC=10} & Random & 45.72$\pm{0.78}$ & 90.05$\pm{0.38}$ & 43.29$\pm{1.08}$ & 21.83$\pm{0.64}$ & 74.42$\pm{0.86}$ & 20.41$\pm{0.87}$ \\
  
    & MAE-Score &  49.04$\pm{0.32}$ & 90.58$\pm{0.32}$ & 46.55$\pm{0.33}$ & \textbf{22.90$\pm{0.28}$} & \textbf{74.95$\pm{0.25}$} & 21.50$\pm{0.38}$ \\
    & CSTA & 49.43$\pm{0.95}$ & \textbf{91.25$\pm{0.63}$} & 47.62$\pm{0.81}$ & 21.96$\pm{0.59}$ & 74.86$\pm{0.92}$ & 21.00$\pm{0.53}$ \\
    & DM-MMD & 49.31$\pm{0.56}$ & 90.27$\pm{0.35}$ &  47.02 $\pm{0.79}$ & 22.41$\pm{0.69}$ & 74.87$\pm{0.81}$ & 20.93$\pm{0.40}$ \\
    & VST-UNet (ours) & 49.72$\pm{0.61}$ & 90.96$\pm{0.26}$ & \textbf{48.67$\pm{0.58}$} & 21.58$\pm{1.25}$ & 73.69$\pm{0.74}$ & 20.60$\pm{1.57}$ \\

    & TAC-DT (ours) & \textbf{49.79\(\pm{0.68}\)} & 91.11\(\pm{0.37}\) & 48.17\(\pm{0.69}\) & 22.48\(\pm{0.99}\) & 74.85\(\pm{0.64}\) & \textbf{21.57\(\pm{1.11}\)} \\
    \midrule
    VPC=100 & - & 56.22$\pm{0.44}$ & 92.99$\pm{0.15}$ & 54.66$\pm{0.36}$ & 31.05$\pm{0.30}$ & 79.76$\pm{0.28}$ & 29.54$\pm{0.34}$ \\
    \bottomrule
  \end{tabular}
  }
\end{center}
  \caption{Results from different clustering methods on the MiniUCF and HMDB51 datasets.}
  \label{diffcluster_mini_hmdb}
\end{table}

In Table~\ref{diffcluster_mini_hmdb}, we compare our approaches with various clustering methods for selecting representative videos from the synthetic datasets \(S_g\) of MiniUCF and HMDB51 datasets.
The selection methods consists of:
\textbf{(a) Random}, a naive baseline method where videos are selected randomly.
\textbf{(b) MAE-Score}, this method ranks videos using scores obtained from a pretrained VideoMAE model~\cite{wang2023videomae}, and selects videos with the highest scores. However, this method does not explicitly consider the dataset diversity.
\textbf{(c) CSTA}. We embed all videos into 1D feature vectors using the pretrained VideoMAE model~\cite{wang2023videomae}.  The features belonging to the same class are then concatenated into a single representation \(I_c \in \mathbb{R}^{N_c \times D}\), which is fed into the pretrained CSTA model~\cite{son2024csta}, originally designed for frame selection within a video. Based on the output scores of all \(N_c\) videos, the knapsack algorithm is applied to select the most representative videos.
\textbf{(d) DM-MMD}, This approach extends distribution matching (DM)~\cite{zhao2023DM} by integrating k-Means clustering and Maximum Mean Discrepancy (MMD) as a probability distance measure.  A DCGAN~\cite{radford2015unsupervised} is used to generate synthetic features aligned with the clustered embeddings.
\textbf{(e) VST-UNet}, our proposed VST-UNet model.
\textbf{(f) TAC-DT}, our TAC-DT method. 
The comparative evaluation in Table~\ref{diffcluster_mini_hmdb} further demonstrates that our selection approaches yield superior classification performance, validating their effectiveness in video dataset distillation.

Table~\ref{diffclassifiers_mini_hmdb} presents a comparison of two action recognition classifiers, C3D~\cite{tran2015learning} and SwinTransformer~\cite{liu2022video}, on the distilled MiniUCF and HMDB51 datasets.
The results show that C3D outperforms SwinTransformer across all VPC settings on both datasets, as the small distilled datasets is insufficient to fully exploit the capacity of the larger model. However, SwinTransformer also surpasses the previous state-of-the-art, \emph{e.g.}, the accuracy of \(33.21\) vs. \(31.2\) on MiniUCF with VPC \(= 5\). 
This demonstrates that our video selection method is classifier-agnostic and effectively enhances performance across different architectures.

\begin{table}[tb]
\begin{center}
  \resizebox{1\columnwidth}{!}{%
  \begin{tabular}{llcccccc}
    \toprule
    &  & \multicolumn{3}{c}{MiniUCF} & \multicolumn{3}{c}{HMDB51} \\
    \midrule
    & Classifiers & ACC & AUC & F1 Score & ACC & AUC & F1 Score \\
    \midrule
    \multirow{2}{*}{VPC=1} & FRePo\_VD~\cite{wang2024dancing} & 22.0\(\pm{1.0}\) & - & - & 8.6\(\pm{0.5}\) & - & - \\
    & C3D~\cite{tran2015learning} & \textbf{26.28$\pm{1.07}$} & \textbf{80.30$\pm{0.45}$} & \textbf{24.68$\pm{0.36}$} & \textbf{13.59$\pm{0.26}$} & \textbf{65.53$\pm{0.45}$} & \textbf{11.80$\pm{0.37}$} \\
    & Swin~\cite{liu2022video} & 18.93\(\pm{0.85}\) & 77.50\(\pm{0.83}\) & 17.22\(\pm{0.54}\) & 10.20\(\pm{0.90}\) & 63.36\(\pm{0.62}\) & 8.65\(\pm{0.75}\) \\
    
    \midrule
    \multirow{2}{*}{VPC=5} & FRePo\_VD~\cite{wang2024dancing} & 31.2\(\pm{0.7}\) & - & - & 10.3\(\pm{0.6}\) & - & - \\
    & C3D~\cite{tran2015learning} & \textbf{41.81\(\pm{0.99}\)} & \textbf{88.45$\pm{0.54}$} & \textbf{40.07$\pm{0.95}$} & \textbf{18.05$\pm{0.75}$} & \textbf{71.87$\pm{0.78}$} & \textbf{17.15$\pm{0.80}$} \\
    & Swin~\cite{liu2022video} & 33.21\(\pm{0.95}\) & 85.70\(\pm{0.47}\) & 31.90\(\pm{1.05}\) & 14.44\(\pm{0.88}\) & 70.14\(\pm{0.43}\) & 13.64\(\pm{0.96}\) \\
    
    \midrule
    \multirow{2}{*}{VPC=10} & C3D~\cite{tran2015learning} & \textbf{49.72$\pm{0.61}$} & \textbf{90.96$\pm{0.26}$} & \textbf{48.67$\pm{0.58}$} & \textbf{21.42$\pm{0.83}$} & \textbf{74.73$\pm{0.56}$} & \textbf{19.96$\pm{0.76}$} \\ 
    & Swin~\cite{liu2022video} & 39.62\(\pm{1.59}\) & 89.09\(\pm{0.50}\) & 38.72\(\pm{1.64}\) & 17.54\(\pm{0.45}\) & 72.86\(\pm{0.51}\) & 16.84\(\pm{0.35}\) \\
    \bottomrule
  \end{tabular}
  }
\end{center}
  \caption{Comparison of different action classifiers, C3D~\cite{tran2015learning} and SwinTransformer~\cite{liu2022video}, using VST-UNet method.}
  \label{diffclassifiers_mini_hmdb}
\end{table}

We evaluate the impact of different loss functions for training the VST-UNet model on the MiniUCF dataset in Table~\ref{difflosses}, with the number of VPC set to 5. The combination of all three loss functions yields the best performance, with each individual loss contributing a slight improvement.
Table~\ref{computationcost} summarizes the computational cost of our approach. The training of video diffusion models can be considered part of the preparation phase, as the synthetic videos are generated only once and reused throughout the distillation process. The reported values are the total computation time aggregated over all \(5\) runs. Overall, our method requires significantly less time compared to typical video tasks.

\begin{table}[tb]\setlength{\tabcolsep}{10pt}
\begin{center}
  \begin{tabular}{llccc}
    \toprule
    & Losses & ACC & AUC & F1 Score \\
    \midrule
    & CE & 31.77\(\pm{0.50}\) & 82.52\(\pm{0.35}\) & 29.64\(\pm{0.49}\)  \\
    & CE+Div & 40.18\(\pm{0.87}\) & 87.37\(\pm{0.64}\) & 38.12\(\pm{0.91}\) \\
    & CE+Rep & 39.80\(\pm{1.27}\) & 87.80\(\pm{0.99}\) & 37.81\(\pm{1.68}\)  \\
    & CE+Div+Rep & \textbf{41.81\(\pm{0.99}\)} & \textbf{88.45\(\pm{0.54}\)} & \textbf{40.07\(\pm{0.95}\)}  \\
    
    \bottomrule
  \end{tabular}
\end{center}
  \caption{Results of different losses in the VST-UNet method on the MiniUCF dataset. CE is the cross entropy loss, Div is the diversity loss, and Rep is the representative loss.}
  \label{difflosses}
\end{table}

\begin{table}[tb]
\begin{center}
  \resizebox{0.98\columnwidth}{!}{%
  \begin{tabular}{llccccccc}
    \toprule
    && GPU & VPC & UCF101 & MiniUCF & HMDB51 & Kinetics400 & SSv2 \\
    \midrule
    \multirow{2}{*}{Diffusion} & Time & 4 & -- & -- & -- & \(\sim\) 3 days & \(\sim\) 7 days & \(\sim\) 6 days \\
    & Steps & -- & -- & --& -- & 500k & 1M & 1M \\
    \midrule
    \multirow{4}{*}{Distill} & VST-UNet & 1 & 1 & 3.5 h & 1.5 h & 2.5 h  & 20 h & 16 h \\
    &TAC-DT & 1 & 1 & 2.5 h & 1 h & 1.5 h & 17 h & 14.8 h \\
    \cmidrule{2-9}
    &VST-UNet & 1 & 5 & 11.5 h & 6 h & 6 h & 67 h & 33.5 h \\
    &TAC-DT & 1 & 5 & 7 h & 2.5 h & 2.7 h & 38 h & 25.5 h \\
    \bottomrule
  \end{tabular}
  }
\end{center}
  \caption{The computation cost. The time consists of \(5\) runs.}
  \label{computationcost}
\end{table}

\section{Conclusion}

We propose a video dataset distillation framework that effectively reduces data storage and computational costs while maintaining strong classification performance. Our approach leverages a latent diffusion model to generate high-quality synthetic videos, ensuring comprehensive coverage of the original dataset distribution.  
To construct the synthetic dataset, we introduce VST-UNet, a spatio-temporal video selection model, which identifies the most essential videos for training. 
To further enhance computational efficiency, we introduce TAC-DT, a training-free clustering algorithm for video selection.
Our results significantly outperform existing methods. Furthermore, when setting VPC to \(10\), our approach achieves performance comparable to training on the full real dataset, highlighting the effectiveness of our selection strategies in preserving essential information while reducing data volume.

\paragraph{Future Work.}
While we report total distillation time as a measure of computational efficiency, future work will provide a more comprehensive assessment by incorporating additional metrics such as FLOPs and memory usage. These metrics will offer a deeper understanding of the trade-offs between computational cost and performance. We also plan to investigate strategies for improving both performance and training efficiency on large-scale datasets, such as Kinetics400 and Something-Something V2, thereby enhancing the practicality of our method for broader deployment.

\paragraph{Acknowledgments:} High-performance computing resources were provided by the Erlangen National High Performance Computing Center (NHR@FAU) at Friedrich-Alexander-Universität Erlangen-Nürnberg (FAU), under the NHR projects b143dc and b180dc. NHR is funded by federal and Bavarian state authorities, and NHR@FAU hardware is partially funded by the German Research Foundation (DFG) – 440719683. Additional support was  received by the ERC - project MIA-NORMAL 101083647,  DFG 513220538, 512819079, and by the state of Bavaria (HTA). We acknowledge the use of Isambard-AI National AI Research Resource (AIRR)~\cite{mcintosh2024isambard}. Isambard-AI is operated by the University of Bristol and is funded by the UK Government’s DSIT via UKRI; and the Science and Technology Facilities Council [ST/AIRR/I-A-I/1023].
S. Cechnicka is supported by the UKRI Centre for Doctoral Training AI4Health  (EP / S023283/1).


\bibliography{egbib}
\end{document}